\def\year{2019}\relax
\documentclass[letterpaper]{article} 
\usepackage{aaai19}  
\usepackage{times}  
\usepackage{helvet}  
\usepackage{courier}  
\usepackage{url}  
\usepackage{graphicx}  
\usepackage{subfigure}
\usepackage{bbm}
\usepackage{amsthm}
\usepackage{amsmath}
\usepackage{algorithm,algpseudocode}

\newtheorem{theorem}{Theorem}
\newtheorem{definition}[theorem]{Definition}
\begin{document}
%
\title{Efficient Metropolitan Traffic Prediction Based on \\ Graph Recurrent Neural Network}
\author{Xiaoyu Wang$^\dagger$, Cailian Chen$^\dagger$, Yang Min$^\dagger$, Jianping He$^\dagger$, Bo Yang$^\dagger$, Yang Zhang$^\ddag$ \\
$^\dagger$The Dept. of Automation, Shanghai Jiao Tong University, and the Key Laboratory of System Control \\
and Information Processing, Ministry of Education of China, Shanghai, China \\
$^\ddag$Shanghai Transportation Information Center, Shanghai, China \\
\{xxArbiter, cailianchen, purifyang, jphe, bo.yang\}@sjtu.edu.cn, china.zhangyang@foxmail.com}
\nocopyright
\maketitle
\begin{abstract}
Traffic prediction is a fundamental and vital task in Intelligence Transportation System (ITS), but it is very challenging to get high accuracy while containing low computational complexity due to the spatiotemporal characteristics of traffic flow, especially under the metropolitan circumstances. In this work, a new topological framework, called Linkage Network, is proposed to model the road networks and present the propagation patterns of traffic flow. Based on the Linkage Network model, a novel online predictor, named Graph Recurrent Neural Network (GRNN), is designed to learn the propagation patterns in the graph. It could simultaneously predict traffic flow for all road segments based on the information gathered from the whole graph, which thus reduces the computational complexity significantly from $O(nm)$ to $O(n+m)$, while keeping the high accuracy. Moreover, it can also predict the variations of traffic trends. Experiments based on real-world data demonstrate that the proposed method outperforms the existing prediction methods.
\end{abstract}

\section{Introduction}
An accurate traffic prediction in metropolitan circumstance is of great importance to the administration department. Taking Transportation Information Center (TIC) of Shanghai as an example, the high-accuracy traffic prediction helps to control the traffic flow. At the same time, the occurrences of large-scale traffic congestion always imply the gathering of citizens. Thus, traffic prediction also helps to prevent public or traffic accidents from happening \cite{zheng2014urban} through noticing administrators in advance, and the emergency response plans can be deployed promptly.

\begin{figure}[t]
\centering
  \subfigure[Strong seasonal trend] {\label{fig:regular} \includegraphics[width=0.24\textwidth]{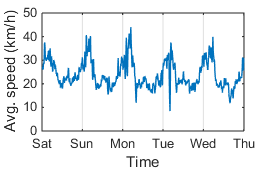}}
  \hspace{-15pt}
  \subfigure[Week seasonal trend] {\label{fig:irregular} \includegraphics[width=0.24\textwidth]{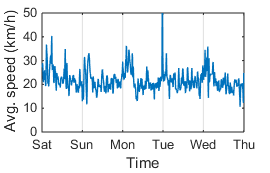}}
\vspace{-0pt}
\caption{Seasonal trend of differen road segments}
\vspace{-0pt}
\end{figure}

There was research \cite{nguyen2017discovering} focusing on such a meaningful problem, but still left with some limitations. Firstly, most of the existing approaches, \cite{lippi2013short-term,fusco2016short-term} for example, consider traffic prediction as a time series problem and solve it with common methods in the disciplines of time series analysis and statistical learning. However, the traffic condition of one road segment is strongly correlated to the others¡¯. Thus, the global information of the whole traffic network is ignored. Secondly, the traffic condition of some segments has an obvious seasonal regularity as the example shown in Figure \ref{fig:regular}, but most segments do not have such characteristics \ref{fig:irregular}. This phenomenon restrains the performance of the methods which only excavate numerical correlations and exacerbates the difficulty of prediction. Thirdly, some approaches \cite{min2011real-time,Zhang2016DNN} introduce additional spatiotemporal data to assist the prediction. They addressed the global information to some extent, but the extra data also leads to copious expenditures on computation. And the existence of the strong coupling in the road system in both time and space indicates that the prediction using local information separately is eventually not equal to the prediction from global to global simultaneously. Hence, traffic condition prediction is still a tough problem remaining to be solved.

In this paper, we propose a novel scheme to handle the limitations mentioned above, which consists of two key parts: the linkage network and the online regressor Graph Recurrent Neural Network (GRNN). First of all, we define the Linkage Network to enrich the properties a graph of the road network can present. Linkage which is newly introduced can include and present the significant property called propagation pattern, which actually shows the internal mechanism of the traffic variation.

After that, GRNN is proposed to mine and learn this propagation pattern and make the prediction \emph{globally and synchronously}. GRNN contains a propagation module to propagate the hidden states along the linkage network just as the traffic flow spreading along the road network. Considering that the propagation of traffic flow directly affects the variation of traffic, GRNN can easily generate the prediction results with the already learned patterns. In conclusion, our contribution can be sum up into four folds\@:
\begin{enumerate}
\item[$\bullet$] Linkage Network is modeled to dislodge the useless redundancy in the traditionally defined road network, and its new element linkage can contain and present the vital feature called propagation pattern, which is the major cause of the traffic variation.
\item[$\bullet$] GRNN, which can absorb the information from the whole graph, is designed to mine and learn the propagation pattern in the linkage network, and it can further generate traffic prediction directly from the features it learned.
\item[$\bullet$] We derive and give the learning algorithm of GRNN, and additionally prove that the computational complexity is lower than traditional approaches.
\item[$\bullet$] We evaluate our scheme using taxi trajectory data of Shanghai. The experiment results demonstrate the advantages of the new scheme we proposed compared with 5 baselines.
\end{enumerate}

\section{Problem Formulation}
In this section, we briefly introduce the traffic prediction problem. Here we give the most commonly used definition of road network firstly.
\begin{definition}\label{d:network}
    \textbf{Road Network.} A traditional road network $G(\mathcal{V},\mathcal{E})$ is defined as a directed graph, where $\mathcal{V}$ is a set of intersections; while $\mathcal{E}$ is a set of road segments. Vertex $v_i$ is defined by the coordinate of intersection $(v_i^{lng}, v_i^{lat})$, which are longitude and latitude respectively. Edge $e_j$ is a segment determined by two endpoints $v^{init}_j, v^{term}_j\in\mathcal{V}$.
\end{definition}

\begin{definition}
    \textbf{Traffic Condition Prediction.} For each road segment $e_i, 1 \le i\le n$ in road network $G$, a time series $\{x^t_i\}$ represents the traffic condition of $e_i$ in each time interval $t$. Traffic condition prediction aims to predict $x^t_i$ from a feature vector $p_i$ using a map $f_i$, and minimize the following error:
    \vspace{-3pt}
    \begin{equation}\label{e:goal}
        L_i := |f_i(\mathbf{p}_i) - x^t_i|.
    \end{equation}
\end{definition}

In most of the traditional approaches, researchers usually aggregate traffic conditions of the to-be-predicted segment in former time steps and other spatiotemporal data as feature vector $\mathbf{p_i}\in\mathbbm{R}^k$, and predict $x^t_i$ of each segment separately, which can be expressed by a map $f:\mathbbm{R}^{k\times n}\mapsto\mathbbm{R}^1$. In our approach, we make the prediction of all segments from global information simultaneously.
\begin{definition}\label{d:globalPred}
    \textbf{Global Traffic Condition Prediction.} In this task, we aim to form a map $f:X\mapsto \hat{\mathbf{x}}^\tau:\mathbbm{R}^{n\times T}\mapsto\mathbbm{R}^n$ maps the prior knowledge of $T$ former steps' conditions of all segments $[\mathbf{x}^{\tau-T}, \ldots, \mathbf{x}^{\tau-1}]\in\mathbbm{R}^{n\times T}$ to the prediction $\hat{\mathbf{x}}^\tau$, and change Equation \ref{e:goal} as follow:
    \begin{equation*}
        L := {\|\hat{\mathbf{x}}^\tau - \mathbf{x}^\tau\|}_q,
    \vspace{-5pt}
    \end{equation*}
    where $\mathbf{x}^\tau$ is the condition vector containing conditions of each road segment in the graph at time $t$, and ${\|\cdot\|}_q$ denotes the loss function measuring the deviation from prediction to actuality, which will be introduced in the following.
\end{definition}

\section{Architecture and the Linkage Network}
We demonstrate the architecture of the whole prediction scheme we proposed and further make a detail explanation of the linkage network in this section.

\subsection{Architecture of the New Scheme}
\begin{figure}[h]
\vspace{-0pt}
\centering
  \includegraphics[width=0.47\textwidth]{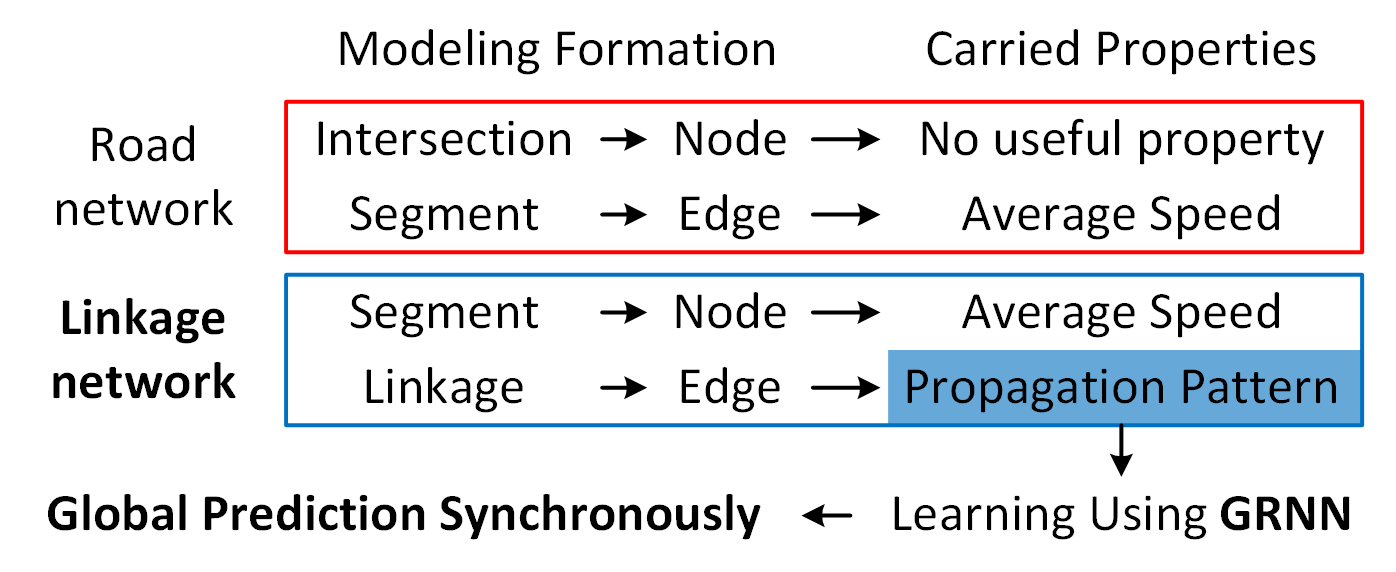}
\vspace{-10pt}
\caption{Architecture of the proposed scheme}
\vspace{-0pt}
\label{fig:flow}
\end{figure}

Our proposed traffic prediction scheme consists of two key model: linkage network and GRNN. The analysis of this architecture is briefly shown in Figure \ref{fig:flow}.

\begin{figure}[h]
\vspace{-0pt}
\centering
  \includegraphics[width=0.3\textwidth]{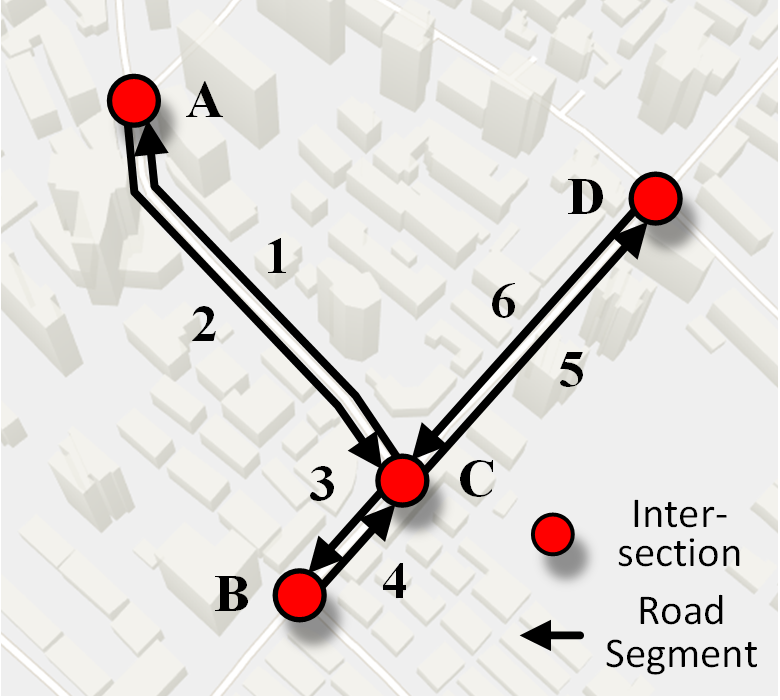}
\vspace{-0pt}
\caption{Example of a subgraph}
\vspace{-0pt}
\label{fig:subgraph}
\end{figure}

In the following, we will always use a subgraph of the whole traffic system, as shown in Figure \ref{fig:subgraph}, as an example.

Road network defined in Definition \ref{d:network} is widely used, which abstracted from actual world intuitively and directly as shown in Figure \ref{fig:subgraph}. Vertexes A to D in the graph represent the intersections which roads converge to, and the directed edges represent unidirectional road segments $1-6$ between intersections.

Road segment has a set of features which can be carried by the vertex in the graph, which usually consists of average speed, number of lanes, length and so on. Meanwhile, two attributes which an intersection has should be analyzed specifically, which are geographic location and the `linkage' among road segments. The former one is meaningless in the topology based research. And we define the latter one, `linkage', as the physical connection between two end to end road segments. For example, at a crossroad, a vehicle has four choices: turn back, left, right and go straight forward. These four choices correspond to four linkages between four downstream road segments and the segment this vehicle currently driving on.

Additionally, we define the `propagation pattern' as the proportion of the vehicles which choose a certain linkage. Back to the illustration of the subgraph in Figure \ref{fig:subgraph}, under the traditional definition, all linkages corresponding to the intersection `C' are coupled together, and propagation patterns have the same coupling problem. As a consequence, although the graph in Definition \ref{d:network} can express the `linkage', the `propagation pattern' a linkage contains can be represented by neither edge nor vertex of it. At the same time, we notice that a large number of vehicles congested on a road segment will eventually spread to downstream segments and increase the burden of them, and the size of traffic flow has a strong relationship with the traffic condition \cite{du2013vanet}. As a consequence, the traffic conditions of downstream road segments are directly influenced by their upstream segments. Thus, the propagation pattern of the urban transportation system is the key to analyze the internal mechanism of the traffic variation. This vital feature has to be decoupled and be expressed by elements in the graph clearly and separately.

Hence, we propose the linkage network to eliminate the useless redundancy of vertexes and compact the propagation pattern in the graph. After that, based on the linkage network, we propose GRNN to mine and learn the propagation pattern from it and further predict traffic in a more efficient way.

\subsection{Linkage Network Modeling}
Here we give the definition of the linkage network:

\begin{definition}\label{d:newnetwork}
    \textbf{Linkage Network.} A linkage network $G(\mathcal{V},\mathcal{E})$ is an unweighted directed graph, where vertexes $\mathcal{V}$ represent the road segments; while directional edges $\mathcal{E}$ denote the linkages between contiguous segments. A directional edge from segment $v_i$ to $v_j$ will be established if and only if the termination intersection of $v_i$ and the initiation intersection of $v_j$ are the same one.
\end{definition}

\begin{figure}[h]
\centering
  \subfigure[Road network] {\label{fig:old} \includegraphics[width=0.22\textwidth]{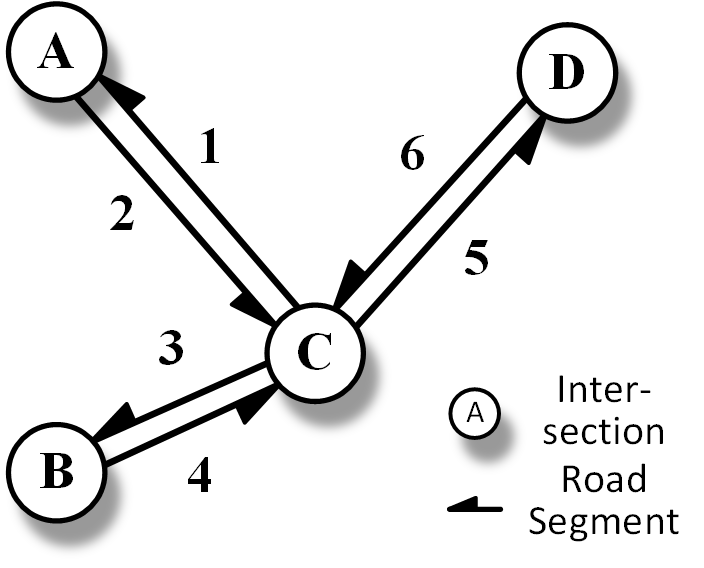}}
  \hspace{-5pt}
  \subfigure[Linkage network] {\label{fig:new} \includegraphics[width=0.24\textwidth]{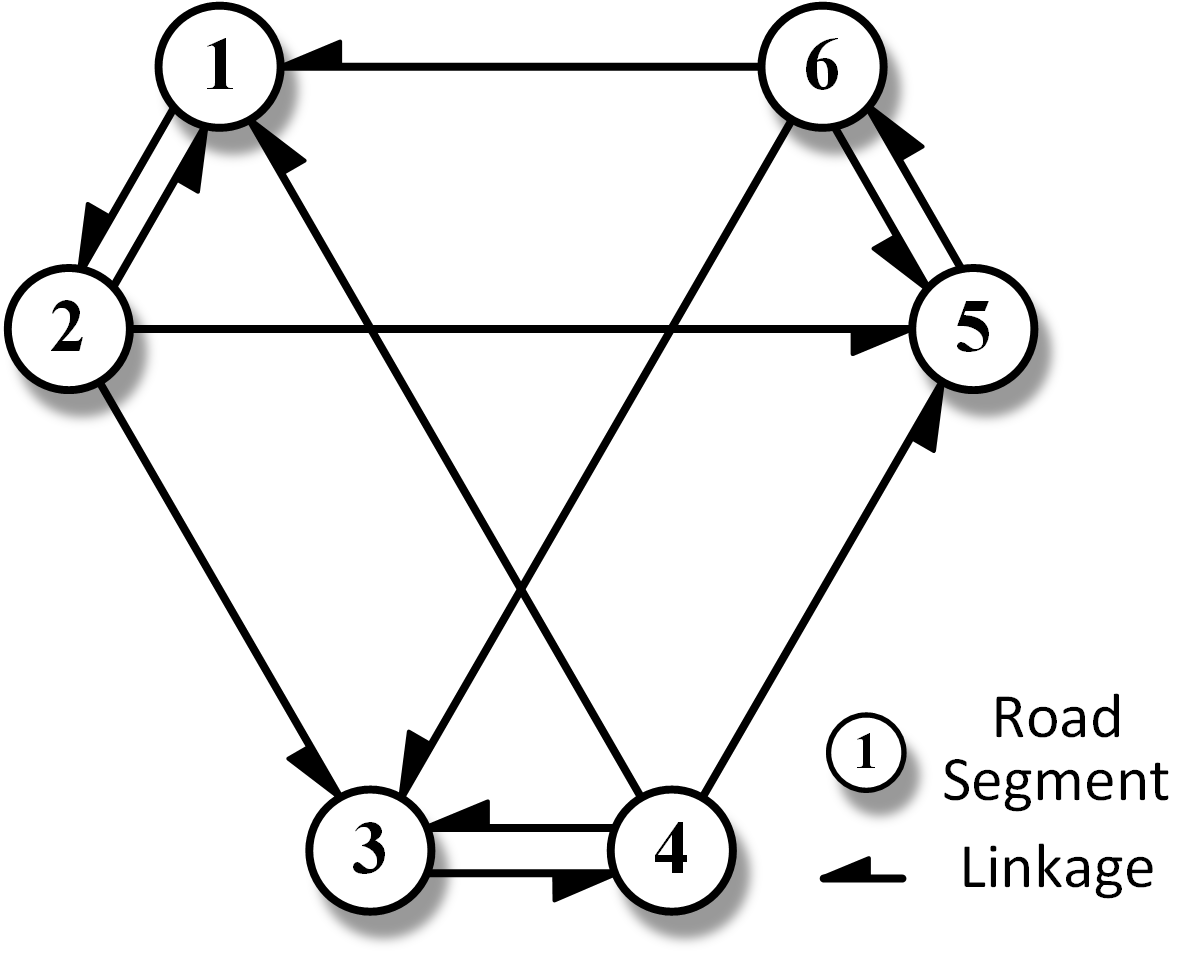}}
\vspace{-0pt}
\caption{Differences between two modeling}
\vspace{-0pt}
\label{fig:network}
\end{figure}

Compared to Figure \ref{fig:old}, Figure \ref{fig:new} illustrates the graph structure of the linkage network. In Definition \ref{d:newnetwork}, the intersection, which carries no useful features, is ignored and the linkage is introduced as the edge of the graph; while the road segment is defined as the vertex. Such a transformation liberates the redundancy of vertexes. Simultaneously, edges now represent linkages containing the traffic propagation patterns, which is significant for the next part of the scheme, the GRNN model, we will introduce. Hence, the linkage network has two main advantages\@:
\begin{enumerate}
\item[$\bullet$] Linkage network can carry is more plentiful information, especially the propagation pattern.
\item[$\bullet$] Only under the definition of linkage network, we can design an algorithm to learn the traffic pattern.
\end{enumerate}

To eliminate the ambiguity, `road network' used in the following represents the road system in the real world, and `linkage network' represents the graph as we defined. At the same time, the new topological structure can be easily transformed from graph defined in Definition \ref{d:network} using the following algorithm we define.

\vspace{-3pt}
\begin{algorithm}
\small
\caption{Graph transformation}
\label{a:graph}
\begin{algorithmic}[1]
    \Require Graph $G(\mathcal{V},\mathcal{E})$ in Definition \ref{d:network}
    \Ensure Adjacency matrix $A$ of $G'(\mathcal{V}',\mathcal{E}')$
    \State $\mathcal{V}' = \{e_i|\,\forall e_i\in\mathcal{E}\}$
    \State Dictionary {$\mathcal{D}\gets\phi$}
    \For {$\forall e_i\in\mathcal{E}$}
        \If {$v^{init}_i == v_j$}
            \State {$\mathcal{D}(v_j) = \mathcal{D}(v_j)\cup e_i$}
        \EndIf
    \EndFor \Comment {Save segments with same initial point}
    \State Chained list {$\mathcal{C}$}
    \For {$\forall v'_i\in\mathcal{V}'$}
        \If {$v^{term}_i == v_j$}
            \State {$\mathcal{C}(v'_i) = \mathcal{D}(v_j)$}
        \EndIf
    \EndFor
    \State Get adjacency matrix $A$ from chained list {$\mathcal{C}$}\\
    \Return {$A$}
\end{algorithmic}
\end{algorithm}

\section{Graph Recurrent Neural Network}
Next, we need an algorithm to complete the global prediction task as defined in Definition \ref{d:globalPred} through the mining and learning from the propagation pattern of traffic flow. Traffic flow is essentially the volume of traffic on each road segment, but the traffic monitoring data we restored in \cite{du2015effective,Wang2018WCSP} is the average speed of vehicles. Fortunately, \cite{du2013vanet} shows that there is a strong relationship between traffic flow and average speed. Thus, we propose the GRNN based on Graph Neural Networks (GNNs) \cite{scarselli2009the} to learn and predict the traffic condition online in an end-to-end form, which means that GRNN will learn the relation between those two metrics and mine the propagation patterns to achieve the goal of global prediction.

The propagation module in GNNs is formed to expand on the time axis of training and propagates hidden states to an equilibrium point to catch the static relationships among vertexes. However, propagation patterns in the transportation system are time-variant, which means that we do not need the propagation process executing too long till stable. Therefore, we compress the propagation module into only one time-step to capture the dynamic relations.

\begin{figure}[t]
\vspace{0pt}
\centering
  \includegraphics[width=0.4\textwidth]{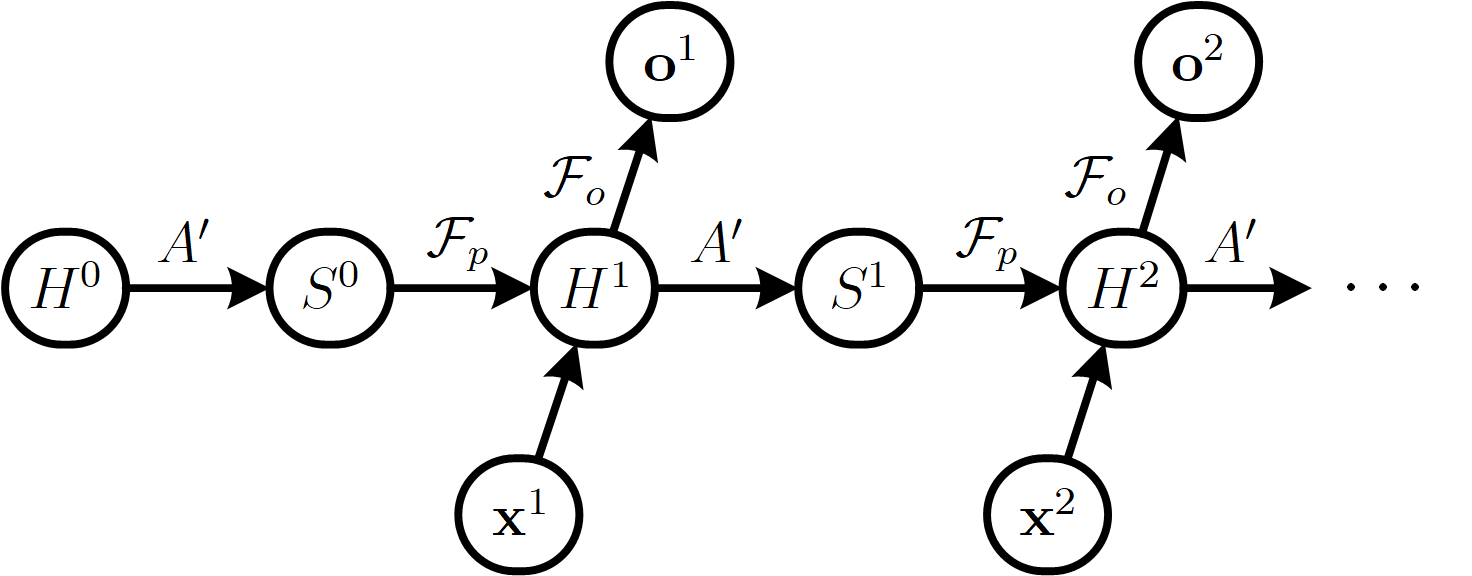}
\caption{Architecture of GRNN}
\vspace{-0pt}
\label{fig:grnn}
\end{figure}

Additionally, the condition of a certain segment is affected by the upstream's conditions not only in the last time step but also in long and short-term history. Thus, we concatenate multiple exactly the same propagation modules end to end to handle the correlations on the time axis of the real world. In other words, the propagation module sends its information back to itself in the next time step. The architecture of GRNN is illustrated in Figure \ref{fig:grnn}, where $\mathcal{F}_p$ and $\mathcal{F}_o$ represent the propagation and output model separately. Under this construction, GRNN also has two major features\@:
\begin{enumerate}
\item[$\bullet$] GRNN becomes a sequence-to-sequence model and overcomes the limitations of GNNs that they have difficulty dealing with streaming data.
\item[$\bullet$] GRNN can learn the propagation pattern represented by the linkage network and predict traffic condition globally and synchronously.
\end{enumerate}

 After all, the Back Propagation Through Time (BPTT) algorithm \cite{Werbos1988Generalization,hochreiter1997long} is utilized to train the whole GRNN.

\subsection{Propagation Module}
\begin{figure}[t]
\centering
  \includegraphics[width=0.30\textwidth]{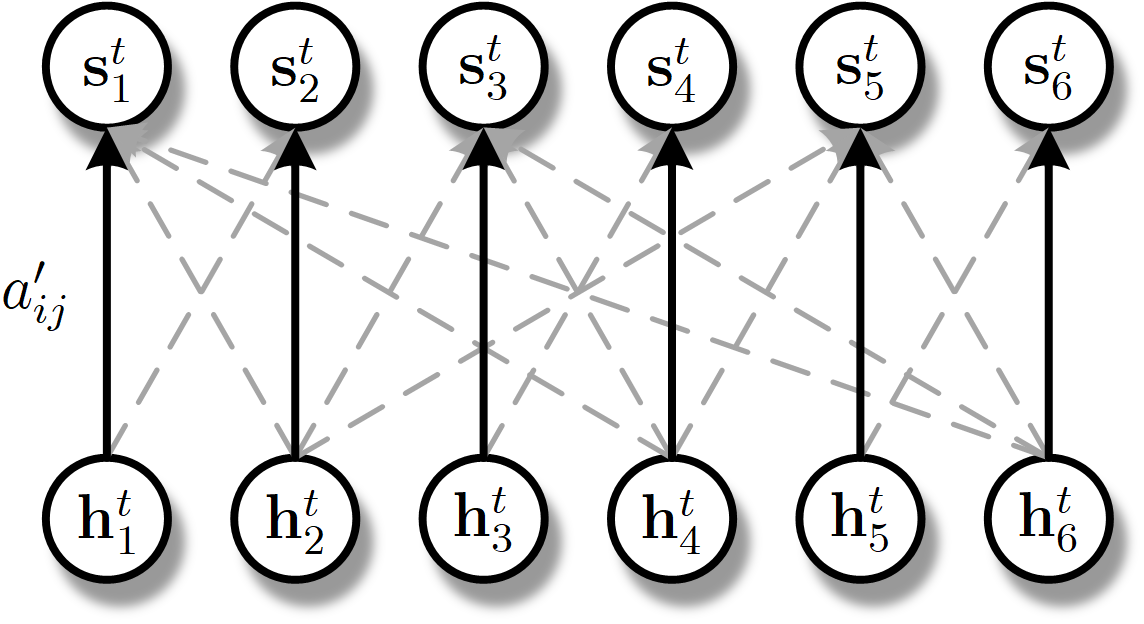}
  \vspace{-5pt}
\caption{Information propagation in the graph, where black arrows denote the self correlation and grey dash lines indicate the mutual correlations.}
\vspace{-0pt}
\label{fig:propa}
\end{figure}

GRNN also use GRU cells in the propagation module like Gated Graph Neural Network (GGNN) \cite{li2016gated} to control the reservation and rejection of information which is gathered from former steps dynamically. In the propagation module of GRNN, the hidden state matrix $H^t = [\mathbf{h}^t_1, \ldots, \mathbf{h}^t_n] \in \mathbbm{R}^{D \times n}$, which represents the propagation patterns here, does not directly relate to the node annotations (time series of traffic condition). $D$ is the dimension of the hidden state of each node. Thus, $H^0$ cannot be initialized through padding a zero matrix on initial node annotations. In GRNN, we randomly initialize $H$ with the normal distribution. Meanwhile, all edges in the linkage network are equal, and the differences of propagation pattern are represented by $H$. Therefore, edges share the same label and all elements in $A$, which controls the propagation direction, are $0$ and $1$. Additionally, since all propagation processes are unidirectional as the definition of the linkage network, $A$ is only a $n \times n$ matrix without any affiliation information. The propagation module of GRNN is formed as follow:
\begin{align}
    S^t             &= H^t \cdot A' \label{e:propa}\\
    Z^{t+1}         &= \sigma\left(W^Z \cdot S^t + U^Z \cdot X^t + B^Z\right) \label{e:z}\\
    R^{t+1}         &= \sigma\left(W^R \cdot S^t + U^R \cdot X^t + B^R\right) \label{e:r}\\
    \tilde{H}^{t+1} &= \mathrm{tanh}\left(W \cdot X^t + U \cdot (R^{t+1} \odot S^t)\right) \label{e:hh}\\
    H^{t+1}         &= (\mathbbm{1}-Z^{t+1}) \odot S^t + Z^{t+1} \odot \tilde{H}^{t+1}, \label{e:h}
\end{align}
where $X^t = [\mathbf{x}^t_1, \ldots, \mathbf{x}^t_n] \in \mathbbm{R}^{d \times n}$ is the input of time $t$. $B^Z, B^R \in \mathbbm{R}^{D \times n}$ are bias matrices. $d$ is the dimension of input feature. $\sigma(x) = 1/(1+e^{-x})$ is the sigmoid function, and $\odot$ is the element-wise multiplication. Considering that the traffic condition of a certain road segment dependents on not only its upstream segments' condition but also its own condition in the last time step. GRNN propagates the states following $A' = \alpha A + I$, where $A \in \mathbbm{R}^{n \times n}$ is the adjacency matrix and $\alpha$ is a hyperparameter which controls the decaying of influence propagation, and $I$ is an equal size identity matrix. Take the subgraph in Figure \ref{fig:subgraph} we used above as an example, the propagation among vertexes is illustrated in Figure \ref{fig:propa}. GRNN propagates information and trains model with new inputs as time goes by, which means that hidden states will eventually contain all information from the whole graph. At the same time, since GRNN learns the propagation pattern dynamically, it can be implemented online. At last, equation \ref{e:z}-\ref{e:h} determine the remembering of the incoming information and the forgetting of old information follow a GRU structure.

\subsection{Output Module}
GNN framework provides a flexible way to handle the output. We can easily get node-level or graph-level outputs from the hidden state matrix with different output models. In our regression task, we focus on the node-level predictions for next time steps. Hence we directly construct a fully connected linear layer for the output module as
\begin{equation}\label{e:o}
    \mathbf{o}^t = \sigma\left(\mathbf{w}^o \cdot H^t + \mathbf{b}^o\right),
\end{equation}
where $\mathbf{o}^t = [o^t_1, \ldots, o^t_n]^T$ are prediction results corresponding to $n$ road segments.

\newcounter{mytempeqncnt}
\begin{figure*}[!t]
\normalsize
\setcounter{mytempeqncnt}{\value{equation}}
\setcounter{equation}{9}
\begin{eqnarray}
    \nabla_{H^t}L &=& \biggl\{ \nabla_{H^{t-1}}L\odot(\mathbbm{1}-Z^t) + {W^Z}^\top\left[\nabla_{H^{t-1}}L\odot(Z^t-Z^t\odot Z^t)\odot(\tilde{H}^t-S^{t-1})\right] \biggr. \nonumber\\
    && + (UW^R)^\top\left[\nabla_{H^{t-1}}L\odot Z^t\odot(\mathbbm{1}-\tilde{H}^t\odot\tilde{H}^t)\odot(R^t-R^t\odot R^t)\odot S^{t-1}\right]  \nonumber\\
    && \biggl. + U^\top\left[\nabla_{H^{t-1}}L\odot Z^t\odot(\mathbbm{1}-\tilde{H}^t\odot\tilde{H}^t)\odot R^t\right] \biggr\}\cdot A' + (\nabla_{\mathbf{o}^{t-1}}L\cdot\mathbf{w}^o)^\top. \label{e:dh}
\end{eqnarray}
\begin{align}
    \nabla_{W}L    =& \textstyle{\sum}_t M \cdot {X^{t-1}}^\top,           &\nabla_{W^Z}L =&\, \textstyle{\sum}_t M^Z \cdot {S^{t-1}}^\top,
    &\nabla_{U^Z}L =& \textstyle{\sum}_t M^Z \cdot {X^{t-1}}^\top,         &\nabla_{B^Z}L =&\, \textstyle{\sum}_t M^Z                       \nonumber\\
    \nabla_{U}L    =& \textstyle{\sum}_t M \cdot (R^t \odot S^{t-1})^\top, &\nabla_{W^R}L =&\, \textstyle{\sum}_t M^R \cdot {S^{t-1}}^\top,
    &\nabla_{U^R}L =& \textstyle{\sum}_t M^R \cdot {X^{t-1}}^\top,         &\nabla_{B^R}L =&\, \textstyle{\sum}_t M^R.  \label{e:gradients}
\end{align}
\setcounter{equation}{\value{mytempeqncnt}}
\hrulefill
\vspace*{-5pt}
\end{figure*}

\subsection{Learning Algorithm}
In GRNN we proposed, information is propagated continuously with the progress of online-training. Hence the learning algorithm of it has to be modified based on BPTT. We formula the BPTT algorithm for GRNN with matrices representation as follows. Firstly, we use the Mean Square Error (MSE) as our loss function:
\begin{equation}\label{e:mse}
    L = \frac{1}{nT} \sum^T_{t=1} \left(\mathbf{x}^t - \mathbf{o}^t\right)^2,
\end{equation}
where $T$ is the time span of propagation, and $\mathbf{x}^t$ is the true values. $T$ can be whether the span of whole historical data or a certain value of hyperparameter to truncate the back-propagation process of deviation to simplify the training process. For any time step $t$, the gradient of $L$ with respect to $\mathbf{o}^t$ is formulated as follow:
\begin{equation}\label{e:do}
    \nabla_{\mathbf{o}^t}L = -\frac{2}{nT} \cdot(\mathbf{x}^t-\mathbf{o}^t).
\end{equation}

For the last time step $T$, gradients of $L$ with respect to weight matrices have the simplest forms, for example $\nabla_{H^T}L=((\mathbf{o}^t-\mathbf{o}^t\odot\mathbf{o}^t)\cdot\nabla_{\mathbf{o}^T}L\cdot\mathbf{w}^o)^\top$. For the time steps less than $T$, information of each node will propagate to others. Hence the gradients have different forms, which contain multiple parts of gradients from different nodes from next time steps. The gradients $\nabla_{H^t}L, t<T$ are too sophisticated to express, so we give out the recursion form.


Under the representations of Equation \ref{e:do}, \ref{e:dh}, gradients of $L$ with respect to weight matrices in the time steps less than $T$ can be expressed in Equation \ref{e:gradients}, and:
\begin{align}
    M   =&\,\, \nabla_{H^t}L \odot Z^t \odot \left(\mathbbm{1}-\tilde{H}^t\odot\tilde{H}^t\right)                   \nonumber\\
    M^Z =&\,\, \nabla_{H^t}L \odot \left(Z^t-Z^t\odot Z^t\right) \odot \left(\tilde{H}^t-S^{t-1}\right)             \nonumber\\
    M^R =&\,\, U^\top\cdot \biggl(\nabla_{H^t}L\odot Z^t\odot\left(\mathbbm{1}-\tilde{H}^t\odot\tilde{H}^t\right)   \nonumber\\
         &\,\, \odot\left(R^t-R^t\odot R^t\right)\biggr) \cdot {S^{t-1}}^\top.                                      \nonumber
\end{align}

From the equations above, we can clearly see correlations among hidden states of all vertexes in different time steps. Finally, the online training and prediction process of GRNN is described in Algorithm \ref{a:grnn}.

\vspace{-2pt}
\begin{algorithm}
\small
\caption{GRNN Online Training and Prediction}
\label{a:grnn}
\begin{algorithmic}[1]
    \Require Historical Data: $\mathbf{x}^{t-T+1},\ldots,\mathbf{x}^t$; Old model
    \Ensure Predictions: $\mathbf{o}^{t+1}$; Updated model
    \State Load {$H^{t-T+1}$} from model trained in last time steps
    \State Predict {$\mathbf{o}^{t+1}$} with {$H^{t-T+1}$} through Equation \ref{e:propa} to \ref{e:h}
    \For{Each iteration epoch $i$}
        \State Propagate forward through Equation \ref{e:propa} to \ref{e:h}
        \State Calculate loss through Equation \ref{e:mse}
        \State Update gradients through Equation \ref{e:dh} and \ref{e:gradients}
    \EndFor \\
    \Return Prediction $\mathbf{o}^{t+1}$ and the up-to-date model
\end{algorithmic}
\end{algorithm}

\subsection{Computational Complexity}
Here we briefly calculate and compare the computational complexity of GRNN with the traditional single-segment prediction methods.

\noindent\textbf{Space Complexity.} Space complexity, or the usage of computer memory in other words, mainly dependent on the magnitude of parameters to be learned in a model. If we use traditional predictors to predict the traffic conditions of total $n$ road segments in the whole city, there are $n$ models to be trained separately. For each model, memory usage of its weight matrices is $O(D^2)$. Here we replace $D^2$ with $m$, which represents the size of the model to be trained. As a result, the complexity of predictors is $O(nm)$. As for GRNN, memory usage of weight matrices is also $O(m)$, while of hidden states is $O(n)$. Since GRNN shares weight matrices between all nodes, the space complexity of these two parts can simply add together: $O(n+m)$, which is far less than $O(nm)$. Considering the huge amount of road segments ($n=65,836$) a metropolis such as Shanghai has, such a reduction is very meaningful.

\noindent\textbf{Time Complexity.} The comparison of time complexity is more sophisticated. We start with the traditional models as well. Here we suppose that the time complexity in one model is $O(m)$ for simplicity. Thus, the complexity of $n$ models is $O(nm)$ obviously. Back to GRNN, we have to give a more elaborate explanation of Equation \ref{e:dh} and \ref{e:gradients}. The formulas given in those two equations are the simplest for in matrices representation. If we split each of them into vector form, each gradient actually has to be updated $n$ times, whose complexity is similar to update $n$ models one time. However, GRNN can update weights through matrix operations, things have changed. In short, GRNN update parameters only one time in each step. Notice that the size of matrix $H$ is corresponding to $n$, so time complexity of GRNN is far less than $O(nm)$ but slightly larger than $O(m)$, which corresponds to the time complexity of matrix operations of CPU or GPU. Since we can not give out the certain formula of time complexity, a numerical comparison will be presented in next section.

\section{Experiments}

\subsection{Datasets and Settings}
\begin{figure}[t]
\vspace{0pt}
\centering
  \includegraphics[width=0.45\textwidth]{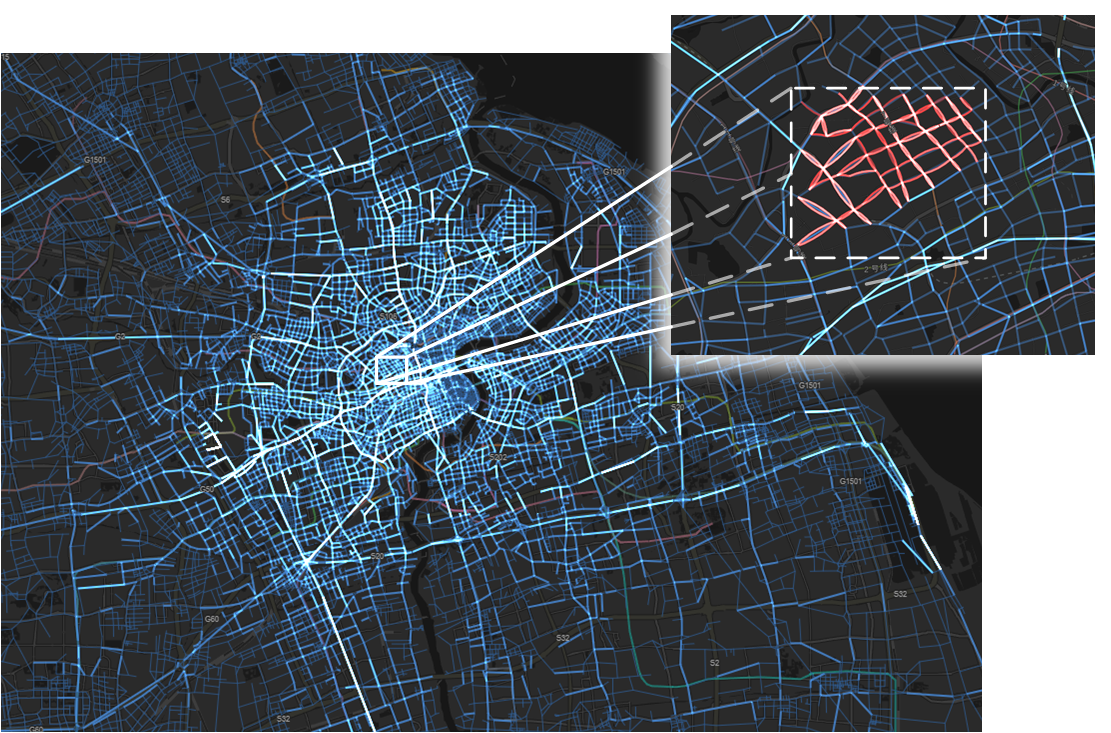}
  \vspace{-5pt}
\caption{Distribution of raw taxi trajectory data, the attached map on the right side show the subgraph we chose, and lighter the segment is, more samples it has.}
\vspace{-5pt}
\label{fig:distri}
\end{figure}

\textbf{Datasets.} Raw taxi trajectory dataset we use in this research is obtained from TIC Shanghai, the distribution of samplings are illustrated in Figure \ref{fig:distri}. To be specific, $310$ GB data are gathered from $13,573$ taxis from Apr. 1, 2015 to Apr. 30, 2015 and a city-scale road network contains $65,836$ road segments. Each taxi reports the GPS report every 10 seconds. The raw trajectory data include the ID, geographical position, upload time stamp, carrying state, speed, the orientation of the vehicle and so on. We mined and restored the traffic conditions of all segments in that time span in our previous work \cite{Wang2018WCSP}, and set the time interval to $10$ minutes. Unfortunately, samples from most of the segments are too sparse, in other words, we only have a set of segments with entire time series of traffic conditions. Thus, we select a connected subgraph with $156$ vertexes as shown in the attached graph on the right side of Figure \ref{fig:distri} with highest sampling density as our test bed where all following experiments will be executed. To be noticed, all the raw data are private, but the processed testbed is available on GitHub, together with the codes of the proposed scheme and tests: https://github.com/xxArbiter/grnn.

\noindent\textbf{Baselines.} We compare our scheme with the following five baselines:
\begin{enumerate}
\item[$\bullet$] \textbf{HA}: Historical average predicts traffic condition by the average value of conditions of the corresponding time interval of each passing day.
\item[$\bullet$] \textbf{ARIMA}
\item[$\bullet$] \textbf{GBRT}: Gradient Boosted Regression Tree is a efficient ensemble model.
\item[$\bullet$] \textbf{SVR}: Support Vector Regression.
\item[$\bullet$] \textbf{GGNN}: GGNN is slightly modified to fit the structure of data in the testbed.
\end{enumerate}

\noindent\textbf{Hyperparameters.} We use PyTorch \cite{pytorch} to construct our model, build ARIMA using Pyramid \cite{pyramid}. SVR and GBRT are implemented with scikit-learn. We select $75\%$ of data for training and the remaining $25\%$ for validation. For SVR and GBDT, we set the dimension of input to $144$, and each GBDT has $200$ trees with up to $7$ layers. In addition, we modify GGNN \cite{li2016gated} slightly to fit our data. Firstly, we add a sigmoid activation function to its output module since the regression task we are facing. Then we put all data of $144$ steps into initial node annotation and set the dimension of the hidden state to $200$ since it has to bigger than the size of annotation. Thus, the initial node representation of each node is $[\mathbf{a}^\top, \mathbf{0}^\top]^\top, \mathbf{a}\in\mathbbm{R}^{144}, \mathbf{0}\in\mathbbm{R}^{56}$. Our GRNN can learn online, which means that it will learn and update itself, and produce the next prediction each time a new set of data is coming, so we truncate the backpropagation process with $T\in\{144, 576, 1008\}$ time steps. Experiments will also be executed to show the effect of the other two extra hyperparameters: the dimension of the hidden state $D$ and iteration epochs $i$.

\noindent\textbf{Evaluation metrics.} We evaluate our method by MSE, which shares the definition in Equation \ref{e:mse}, and Variance of Deviation (VD). Additionally, Running Time (RT) will be used to judge the computational complexity of GRNN.
\setcounter{equation}{11}
\begin{eqnarray}
    MSE \!\!\!&=&\!\!\! \frac{1}{nT}\sum^n_{i=1} \sum^T_{t=1} (e^t_i)^2 \\
    VD  \!\!\!&=&\!\!\! \frac{1}{nT}\sum^n_{i=1} \sum^T_{t=1} (e^t_i - \overline{e})^2,
\end{eqnarray}
where $e^t_i = x^t_i - o^t_i$ is the prediction error of each vertex, each time interval, and $\overline{e}$ is the mean of all errors. VD is defined to measure the dispersity of prediction deviation. A higher VD means that the model cannot track the true value promptly, in other words, it cannot forecast the peaks. The usage of VD will be explained in details in the following.

\subsection{Results of Experiments}
We give the comparison between our model and baselines with metrics MSE and VD as shown in Table \ref{t:comp}.

\begin{table}[h]
\small
	\centering
    \vspace{-10pt}
	\caption{Comparison among different methods}
	\begin{tabular}{l | c | c}
            \hline
		Model                       &   MSE         &   VD          \\
            \hline
		HA                          &   22.920      &   22.878      \\
		ARIMA                       &   14.750      &   14.396      \\
        SVR                         &   22.230      &   18.135      \\
        GBDT                        &   7.082       &   7.076       \\
        Modified GGNN               &   7.091       &   7.051       \\
            \hline
        \textbf{GRNNs}[ours]        &               &               \\
        GRNN-144T-32D-2i            &   6.543       &   6.537       \\
        GRNN-144T-32D-10i           &   5.576       &   5.576       \\
        GRNN-576T-32D-10i           & \textbf{4.540}& \textbf{4.539}\\
        GRNN-576T-64D-10i           &   4.779       &   4.779       \\
        GRNN-1008T-32D-10i          &   4.850       &   4.850       \\
        GRNN-1008T-64D-10i          &   5.405       &   5.405       \\
            \hline
	\end{tabular}
	\label{t:comp}
\end{table}

Results of several versions of GRNNs with different hyperparameters are also listed. It is obvious to see that GRNN-576T-32D-10i has the best performance, but even the smallest GRNN-144T-32D-2i can also outperform traditional time series analysis methods. We make a specific analysis of the influence of different hyperparameters further. Firstly, with the enlargement of the network scale, $i$ has to be increased together since the information to be learned is more detailed, but $i$ has an upper limit. For example, GRNN-144T-32D-100i will always diverge while a certain time. This indicates that too many iterations will converge the model to a wrong equilibrium. Simultaneously, $T$ has to be scaled together with $D$ from the same aspect of the more detailed information to be learned. Notably, the modified GGNN also have a relatively well performance and low complexity, but our algorithm beats it with a smaller size.

\begin{figure}[t]
\centering
  \subfigure[Prediction of GRNN] {\label{fig:resgrnn} \includegraphics[width=0.23\textwidth]{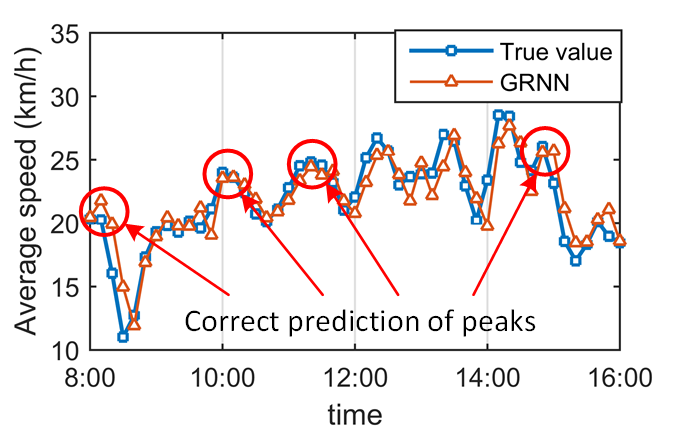}}
  \hspace{-5pt}
  \subfigure[Prediction of GBDT] {\label{fig:resgbdt} \includegraphics[width=0.23\textwidth]{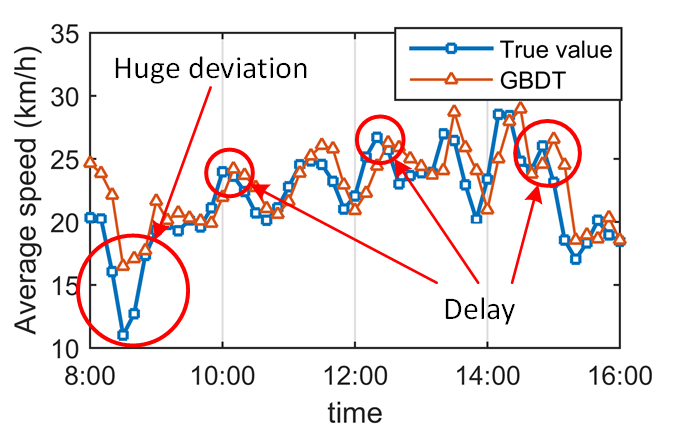}}
\caption{Performance of GRNN and GBDT}
\vspace{0pt}
\label{fig:res}
\end{figure}

Additionally, we chose a random road network to show the results in details. Here we compare GRNN-576T-32D-10i and GBDT, which achieves the highest accuracy among traditional methods. Predictions from 8:00 to 16:00, 27th April 2015, are shown in Figure \ref{fig:res}. Except for the higher accuracy we discussed above, GRNN can track the ground truth more promptly. Specifically, GRNN predicts the peaks correctly at the positions marked by the red circles in Figure \ref{fig:resgrnn}. Meanwhile, the peaks predicted by GBDT always have phase differences, in other words, results of GBDT delay the true values. This phenomenon can also be indicated by the metric VD we defined above.

\noindent\textbf{Numerical explanation of time complexity.} Here we compare the running time of GRNN with different $n$ using a relatively small model to give out an intuitive cognition of time complexity of GRNN. We execute experiments with $1, 10, 156$ nodes separately, and chose the mode GRNN-144T-32D-10i-0.01Lr. Comparison results are briefly shown in Table \ref{t:time}, where the MSE and VD are the metrics of a certain road segment, rather than the whole graph. Through these experiments, we can clearly see that the running time remains almost unchanged with the growing of graph size. These results indicate the conclusion, that the time complexity of GRNN is far less than $O(nm)$ and approximate to $O(m)$, we made in the last Section. Additionally, it is clear to see that the accuracy of prediction will increase together with the enlargement of the scale of the subgraph, which verifies the inference that the propagation patterns GRNN learned from graph contribute to the effectiveness of prediction. And the reduction of accuracy when $n=156$ can also be explained by the superfluous information to be learned, which can be relieved by expansion of the network.

\vspace{-10pt}
\begin{table}[h]
\small
	\centering
	\caption{Numerical experiment of time complexity}
	\begin{tabular}{c | c  c  c}
            \hline
		$n$               &   1           &   10              &   156         \\
            \hline
		MSE of road \#1   &   5.4053      &   \textbf{3.482}  &   4.072       \\
		VD of road \#1    &   5.4050      &   \textbf{3.478}  &   3.715       \\
        RT/min            &   178.31      &   176.40          &   181.57      \\
            \hline
	\end{tabular}
	\label{t:time}
\end{table}

\section{Related Work}
\textbf{Traffic condition prediction.} There are many previous works \cite{Chen2012The} considering the traffic condition as a time series and predicting for different segments separately through time series analysis, like Auto-Regressive Moving Average (ARMA) based algorithms (ARIMA, SARIMA). Additionally, some research \cite{Oh2016Improvement,Hu2016Crowdsourcing} uses the methods of statistical learning such as Bayesian Network (BN), SVR and GBDT, and adds extra information to assist the training. \cite{fusco2016short-term} compares those methods and shows their similar performances. In these approaches, the strong spatiotemporal couplings, which exist in metropolitan circumstance particularly, lead to the dilemma of choices between the computation complexity and the sufficiency of input information.

\cite{Li2015Robust} tries to mine the relationship between consecutive monitoring stations on the highway to predict traffic condition. It is an improvement but the correlation between stations is very intuitive. \cite{zhang2016deep,polson2017deep} delimit urban area into grids and predict the flow of citizen with deep learning algorithms like Convolutional Neural Network (CNN) and Residual Network (ResNet). Although these approaches can learn globally, the action of gridding has already broken the topological structure of the road network. \cite{liang2017inferring} infers the cascading pattern of traffic flow with tree searching and forecasts the congestion further, but it is eventually a local learning method.

\noindent\textbf{Learning of graph-structured data.} Few frontier investigations focus on learning from graph-structured data. \cite{scarselli2009the} earliest proposes the framework of GNN to excavate the relationship in graph-structured data. \cite{Shahsavari2015Short} utilizes it in traffic prediction task and shows its effectiveness. \cite{li2016gated} further expands GNN with GRU cell to simplify the propagation process. Additionally, some works develop another framework call Graph Convolutional Network (GCN) to resolve the graph-structure puzzle in a different way. \cite{niepert2016learning} proposes an application of convolution kernel in the graph domain. Meanwhile, \cite{defferrard2016convolutional,seo2017structured,hamilton2017inductive} establish a various implementation in the frequency domain.

\section{Conclusion and Future Work}
We model a new topological structure for the road network in metropolitan circumstance to remove the useless redundancy and represent more plentiful information and characteristic which cannot be carried by old definition. Further, we propose a novel network GRNN to mine the potential propagation patterns of traffic flow in the redefined graph and achieve the final object of global traffic condition prediction. The outstanding effectiveness of GRNN is shown in experiments and the high-efficiency is proved in the analysis of computational complexity.

In the future, we will expand GRNN with more additional information to achieve higher performance, and explore more application scenarios where data are driven by potential propagation behaviors, economic system and stocks for example. Moreover, we will visualize the patterns GRNN learned from graph-structured data.

\bibliographystyle{aaai}
\bibliography{aaai2019}
\end{document}